\documentclass[runningheads]{llncs}
\usepackage{graphicx}
\usepackage{subfigure}
\usepackage{floatrow}
\usepackage{multirow}
\usepackage{amsmath,amssymb} 
\usepackage{color}

\begin{document}
\title{Skeleton-Based Action Recognition with Spatial Reasoning and Temporal Stack Learning}

\titlerunning{Skeleton-Based Action Recognition}
%
\author{Chenyang Si\inst{1,3} \and
Ya Jing\inst{1,3} \and
Wei Wang\inst{1,3}\thanks{Corresponding Author: Wei Wang} \and
Liang Wang\inst{1,2,3} \and
Tieniu Tan\inst{1,2,3} }
%
\authorrunning{Chenyang Si, Ya Jing, Wei Wang, Liang Wang, and Tieniu Tan}
%

\institute{Center for Research on Intelligent Perception and Computing (CRIPAC),\\
National Laboratory of Pattern Recognition (NLPR) \and
Center for Excellence in Brain Science and Intelligence Technology (CEBSIT), \\
Institute of Automation, Chinese Academy of Sciences (CASIA) \and
University of Chinese Academy of Sciences (UCAS) \\
\email{\{chenyang.si, ya.jing\}@cripac.ia.ac.cn, \{wangwei, wangliang, tnt\}@nlpr.ia.ac.cn}}
\maketitle              
\begin{abstract}
Skeleton-based action recognition has made great progress recently, but many problems still remain unsolved. For example, the representations of skeleton sequences captured by most of the previous methods lack spatial structure information and detailed temporal dynamics features. In this paper, we propose a novel model with spatial reasoning and temporal stack learning (SR-TSL) for skeleton-based action recognition, which consists of a spatial reasoning network (SRN) and a temporal stack learning network (TSLN). The SRN can capture the high-level spatial structural information within each frame by a residual graph neural network, while the TSLN can model the detailed temporal dynamics of skeleton sequences by a composition of multiple skip-clip LSTMs. During training, we propose a clip-based incremental loss to optimize the model. We perform extensive experiments on the SYSU 3D Human-Object Interaction dataset and NTU RGB+D dataset and verify the effectiveness of each network of our model. The comparison results illustrate that our approach achieves much better results than the state-of-the-art methods.

\end{abstract}
\section{Introduction}

Human action recognition is an important and challenging problem in computer vision research. It plays an important role in many applications, such as intelligent video surveillance, sports analysis and video retrieval. Human action recognition can also help robots to have a better understanding of human behaviors, thus robots can interact with people much better \cite{Poppe2010survey,Weinland2011survey,Aggarwal2011Human}.

Recently, there have existed many approaches to recognize human actions, the input data type of which can be grossly divided into two categories: RGB videos \cite{Simonyan2014Two-stream} and 3D skeleton sequences \cite{Du2015Hierarchical}. For RGB videos, spatial appearance and temporal optical flow generally are applied to model the motion dynamics. However, the spatial appearance only contains 2D information that is hard to capture all the motion information, and the optical flow generally needs high computing costs. Compared to RGB videos, Johansson et al.~\cite{johansson1973visual} have explained that 3D skeleton sequences can effectively represent the dynamics of human actions. Furthermore, the skeleton sequences can be obtained by the Microsoft Kinect \cite{zhang2012microsoft} and the advanced human pose estimation algorithms \cite{cao2017realtime}. Over the years, skeleton-based human action recognition has attracted more and more attention \cite{aggarwal2014human,Du2015Hierarchical,Song2017Attention}. In this paper, we focus on recognizing human actions from 3D skeleton sequences.

For sequential data, recurrent neural networks (RNNs) perform a strong power in learning the temporal dependencies. There has been a lot of work successfully applying RNNs for skeleton-based action recognition. Hierarchical RNN \cite{Du2015Hierarchical} is proposed to learn motion representations from skeleton sequences. Shahroudy et al.~\cite{Shahroudy2016NTU} introduce a part-aware LSTM network to further improve the performance of the LSTM framework. To model the discriminative features, a spatial-temporal attention model \cite{Song2017Attention} based on LSTM is proposed to focus on discriminative joints and pay different attentions to different frames. Despite the great improvement in performance, there exist two urgent problems to be solved. First, human behavior is accomplished in coordination with each part of the body. For example, walking requires legs to walk, and it also needs the swing of arms to coordinate the body balance.  It is very difficult to capture the high-level spatial structural information within each frame if directly feeding the concatenation of all body joints into networks. Second, these methods utilize RNNs to directly model the overall temporal dynamics of skeleton sequences. The hidden representation of the final RNN is used to recognize the actions. For long-term sequences, the last hidden representation cannot completely contain the detailed temporal dynamics of sequences.

\begin{figure}[!t]
\centering
\includegraphics[width=1.\linewidth,height=0.38\linewidth]{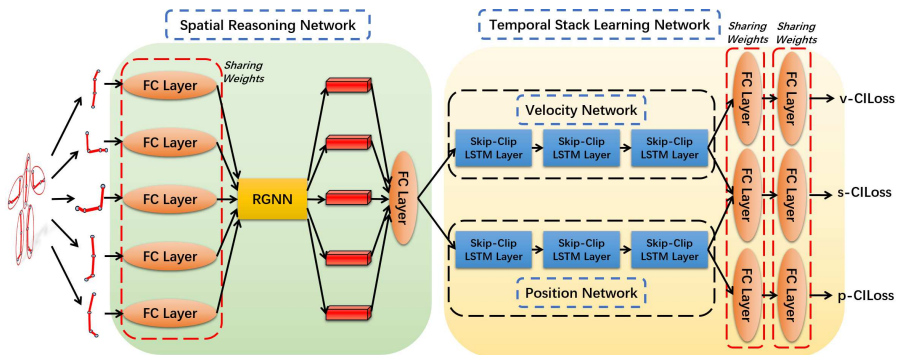}
\caption{The overall pipeline of our model which contains a spatial reasoning network and a temporal stack learning network. In the spatial reasoning network, a residual graph neural network (RGNN) is used to capture the high-level spatial structural information between the different body parts. The temporal stack learning network can model the detailed temporal dynamics for skeleton sequence. During training, the proposed model is efficiently optimized with the clip-based incremental losses (CIloss) }
\label{model_pipeline}
\end{figure}

In this paper, we propose a novel model with spatial reasoning and temporal stack learning (SR-TSL) for this task, which can effectively solve the above challenges. Fig.~\ref{model_pipeline} shows the overall pipeline of our model that contains a spatial reasoning network (SRN) and a temporal stack learning network (TSLN). First, we propose a spatial reasoning network to capture the high-level spatial structural features within each frame. The body can be decomposed into different parts, e.g. two arms, two legs and one trunk. The concatenation of joints of each part is transformed into individual spatial feature with a linear layer. These individual spatial features of body parts are fed into a residual graph neural network(RGNN) to capture the high-level structural features between the different body parts, where each node corresponds to a body part. Second, we propose a temporal stack learning network to model the detailed temporal dynamics of the sequences, which consists of three skip-clip LSTMs. For a long-term sequence, it is divided into multiple clips. The short-term temporal information of each clip is modeled with an LSTM layer shared among the clips in a skip-clip LSTM layer. When feeding a clip into shared LSTM, the initial hidden of shared LSTM is initialized with the sum of the final hidden state of all previous clips, which can inherit previous dynamics to maintain the dependency between clips. We propose a clip-based incremental loss to further improve the ability of stack learning. Therefore, our model can also effectively solve the problem of long-term sequence optimization. Experimental results show that the proposed SR-TSL speeds up the model convergence and improve the performance.

The main contributions of this paper are summarized as follows:
\begin{enumerate}
  \item We propose a spatial reasoning network for each skeleton frame, which can effectively capture the high-level spatial structural information between the different body parts using a  residual graph neural network.
  \item We propose a temporal stack learning network to model the detailed temporal dynamics of skeleton sequences by a composition of multiple skip-clip LSTMs.
  \item The proposed clip-based incremental loss further improves the ability of temporal stack learning, which can effectively speed up convergence and obviously improve the performance.
  \item Our method obtains the state-of-the-art results on the SYSU 3D Human-Object Interaction dataset and NTU RGB+D dataset.
\end{enumerate}

\section{Related Work}
\label{Related Work}
In this section, we briefly review the existing literature that closely relates to the proposed method.

\textbf{\emph{Skeleton based action recognition}} \hspace{3mm}
There have been amounts of work proposed for skeleton-based action recognition, which can be divided into two classes. The first class is to focus on designing handcrafted features to represent the information of skeleton motion. Wang et al.~\cite{Wang2012Mining} exploit a new feature called local occupancy pattern, which can be treated as the depth appearance of joints, and propose an actionlet ensemble model to represent each action. Hussein et al.~\cite{Hussein2013Human} use the covariance matrix for skeleton joint locations over time as a discriminative descriptor for a sequence. Vemulapalli et al.~\cite{Raviteja2014Human} utilize rotations and translations to represent the 3D geometric relationships of body parts in Lie group.

The second class is to use deep neural networks to recognize human actions. \cite{Ke2017A,Kim2017Interpretable} exploit the Convolutional Neural Networks (CNNs) for skeleton-based action recognition. Recently, most of methods utilize the Recurrent Neural Networks (RNNs) for this task. Du et al.~\cite{Du2015Hierarchical} first propose an end-to-end hierarchical RNN for skeleton-based action recognition. Zhu et al.~\cite{Zhu2016Co-Occurrence} design a fully connected deep LSTM network with a regularization scheme to learn the co-occurrence features of skeleton joints. An end-to-end spatial and temporal attention model \cite{Song2017Attention} learns to selectively focus on discriminative joints of the skeleton within each frame of the inputs and pays different levels of attention to the outputs of different frames. Zhang et al.~\cite{Zhang2017View} exploit a view adaptive model with LSTM architecture, which enables the network to adapt to the most suitable observation viewpoints from end to end. A two-stream RNN architecture is proposed to model both temporal dynamics and spatial configurations for skeleton-based action recognition in \cite{Wang2017Modeling}. The most similar work to ours is \cite{Inwoong2017Ensemble} which proposes an ensemble temporal sliding LSTM (TS-LSTM) networks for skeleton-based action recognition. They utilize an ensemble of multi-term temporal sliding LSTM networks to capture short-term, medium-term, long-term temporal dependencies and even spatial skeleton pose dependency. In this paper, we design a spatial reasoning network and temporal stack learning network, which can capture the high-level spatial structural information and the detailed temporal dynamics of skeleton sequences, separately.

\textbf{\emph{Graph neural networks}} \hspace{3mm}
Recently, more and more works have used the graph neural networks (GNNs) to the graph-structured data, which can be categorized into two broad classes. The first class is to apply Convolutional Neural Networks (CNNs) to graph, which improves the traditional convolution network on graph. \cite{henaff2015deep,NIPS2015_5954} utilize the CNNs in the spectral domain relying on the graph Laplacian. \cite{Bruna2014Spectral,niepert2016learning} apply the convolution directly on the graph nodes and their neighbors, which construct the graph filters on the spatial domain. Yan et al.~\cite{Yan2018Spatial} are the first to apply the graph convolutional neural networks for skeleton-based action recognition. The second class is to utilize the recurrent neural networks to every node of the graph. \cite{Scarselli2009graph} proposes to recurrently update the hidden state of each node of the graph. Li et al.~\cite{Li_2017_ICCV} propose a model based on Graph Neural Networks for situation recognition, which can  efficiently capture joint dependencies between roles using neural networks defined on a graph. Qi et.al. \cite{Qi_2017_ICCV} use 3D graph neural networks for RGBD semantic segmentation. In this paper, a residual graph neural network is utilized to model the high-level spatial structural information between different body parts.

\section{Overview}
\label{Overview}

In this section, we briefly review the Graph Neural Networks (GNNs), the Recurrent Neural Networks (RNNs) and Long Short-Term Memory (LSTM), which are utilized in our framework.

\subsection{Graph Neural Network}

Graph Neural Network (GNN) is introduced in \cite{Scarselli2009graph} as a generalization of recursive neural networks, which can deal with a more general class of graphs. The GNNs can be defined as an ordered pair $G$ = \{$V, E$\}, where $V$ is the set of nodes and $E$ is the set of edges. At time step $t$, the hidden state of the $i$-th ($i \in \{ 1,...,\left| V \right| \}$) node is $\vec{s}_i^t$, and the output is $\vec{o}_i^t$. The set of nodes $\Omega_v$ stands for the neighbors of node $v$.

For a GNN, the input vector of each node $v \in V$ is based on the information contained in the neighborhood of node $v$, and the hidden state of each node is updated recurrently. At time step $t$, the received messages of a node are calculated with the hidden states of its neighbors. Then the received messages and previous state $\vec{s}_i^{t-1}$ are utilized to update the hidden state $\vec{s}_i^t$. Finally, the output $\vec{o}_i^t$ is computed with $\vec{s}_i^t$. The GNN formulation at time step $t$ is defined as follows:
\begin{align}
    \label{formu_m} \vec{m}_i^t & = f_m \left( \{ \vec{s}_{\hat i}^{t-1} | {\hat i} \in \{ 1,...,\left| \Omega_{v_i} \right| \} \right) \\
    \label{formu_s} \vec{s}_i^t & = f_s \left( \vec{m}_i^t, \vec{s}_i^{t-1} \right) \\
    \label{formu_0} \vec{o}_i^t & = f_o \left( \vec{s}_i^{t} \right)
\end{align}
where $\vec{m}_i^t$ is the sum of all the messages that the neighbors $\Omega_{v_i}$ send to node $v_i$, $f_m$ is the function to compute the incoming messages, $f_s$ is the function that expresses the state of a node and $f_o$ is the function to produce the output. Similar to RNNs, these functions are the learned neural networks and are shared among different time steps.

\subsection{RNN and LSTM}

Recurrent Neural Networks (RNNs) are the powerful models to capture the dependencies of sequences via cycles in the network of nodes, which are suitable for the sequence tasks. However, there exist two difficult problems of vanishing gradient and exploding gradient when the standard RNN is used for long-term sequences.

The advanced RNN architecture of Long Short-Term Memory (LSTM) is proposed by Hochreiter et al.~\cite{Hochreiter1997lstm}. LSTM neuron contains an input gate, a forget gate, an output gate and a cell, which can promote the ability to learn long-term dependencies.

\section{Model Architecture}
\label{Model Architecture}

In this paper, we propose an effective model for skeleton-based action recognition, which contains a spatial reasoning network and a temporal stack learning network. The overall pipeline of our model is shown in Fig.~\ref{model_pipeline}. In this section, we will introduce these networks in detail.

\subsection{Spatial Reasoning Network}

\begin{figure}[!b]
\centering
    \subfigure[]{
        \label{RGNN:a}
        \includegraphics[width=1.5in]{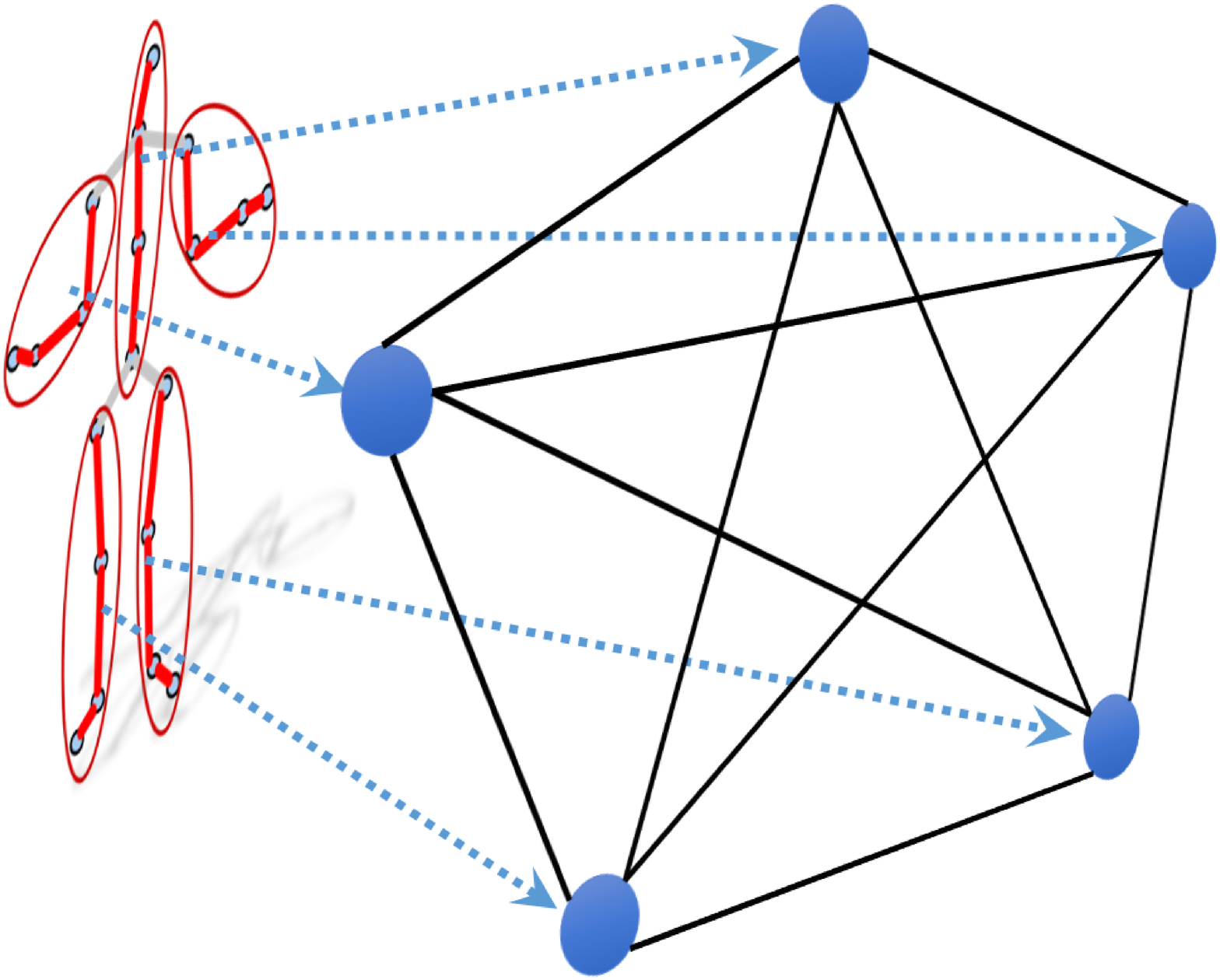}}
    \hspace{8mm}
    \subfigure[]{
        \label{RGNN:b}
        \includegraphics[width=1.65in]{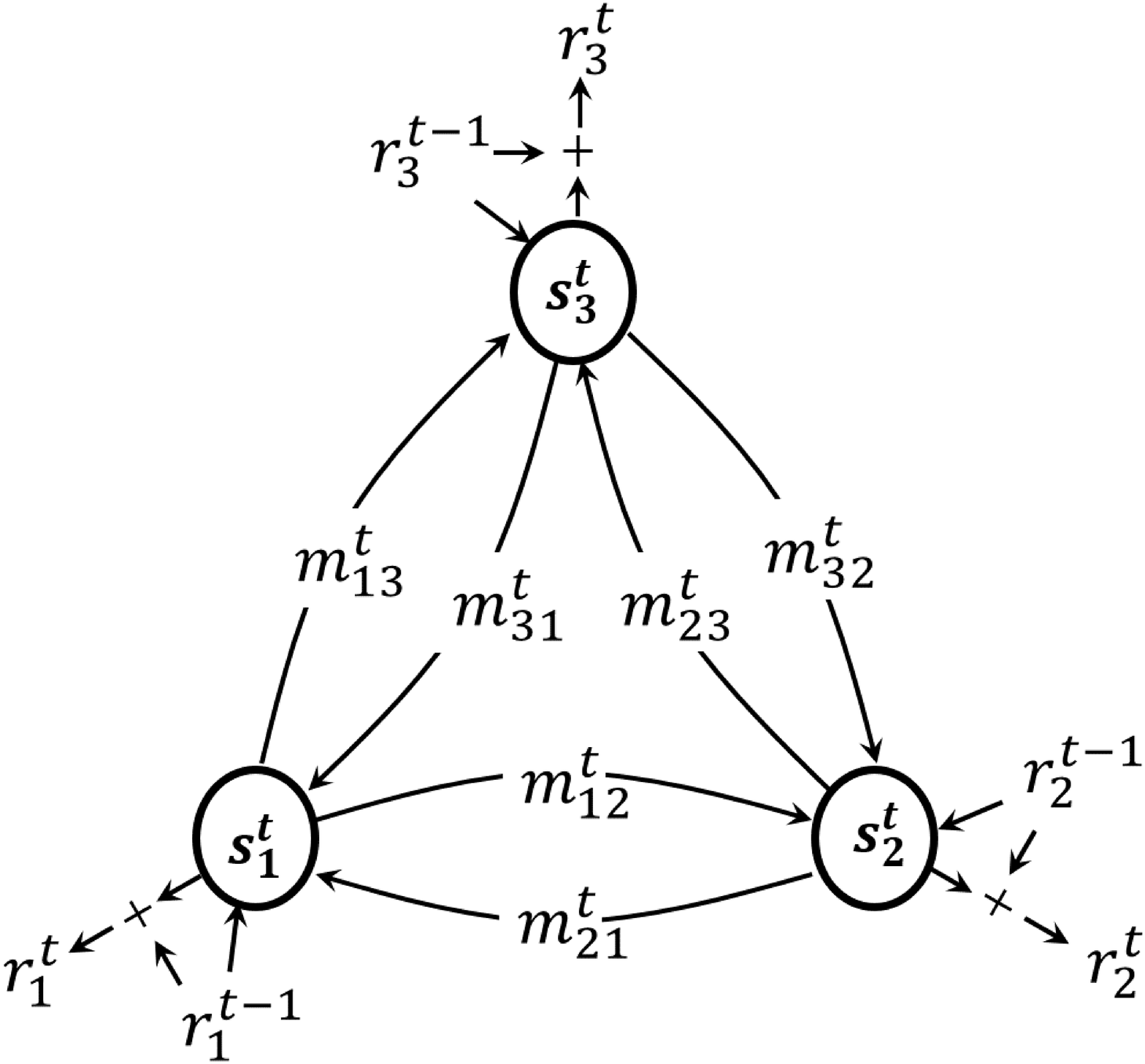}}
\caption{The architecture of residual graph neural network (RGNN). (a) illustrates five human pose parts and a corresponding RGNN. (b) shows the principle of a RGNN with three nodes}
\label{RGNN}
\end{figure}

Rich inherent structures of the human body that are involved in action recognition task, motivate us to design an effective architecture called spatial reasoning network to model the high-level spatial structural information within each frame. According to the general knowledge, the body can be decomposed into $K$ parts, e.g. two arms, two legs and one trunk (shown in Fig. \ref{RGNN:a}), which express the knowledge of human body configuration.

For spatial structures, the spatial reasoning network encodes the coordinate vectors via two steps (see Fig. \ref{model_pipeline}) to capture the high-level spatial features of skeleton structural relationships. First, the preliminary encoding process maps the coordinate vector of each part into the individual part feature $\vec{e}_k$, $k \in \{1,...,K \}$ with a linear layer that is shared among different body parts. Second, all part features $\vec{e}_k$ are fed into the proposed residual graph neural network (RGNN) to model the structural relationships between these body parts. Fig. \ref{RGNN:b} shows a RGNN with three nodes.

For a RGNN, there are $K$ nodes that correspond to the human body parts. At time step $t$, each node has a relation feature vector $\vec{r}_k^t \in R^t$, where $R^t = \{ \vec{r}_1^t,...,\vec{r}_K^T\}$. And $\vec{r}_k^t$ denotes the spatial structural relationships of the part $k$ with other parts. We initialize the $\vec{r}_k^t$ with the individual part feature $\vec{e}_k$, such that $\vec{r}_k^0 = \vec{e}_k$. We use $\vec{m}_{ik}^t$ to denote the received message of node $k$ from node $i$ at time step $t$, where $i \in \{1,...,K \}$. Furthermore, the received messages $\vec{m}_{k}^t$ of node $k$ from all the neighbors $\Omega_{v_k}$ at time step $t$ is defined as follows:
\begin{align}
    \label{rgnn_message} \vec{m}_{k}^t & =  \sum\limits_{i \in \Omega_{v_k}} \vec{m}_{ik}^{t} \nonumber\\
    & = \sum\limits_{i \in \Omega_{v_k}} \vec{W}_m \vec{s}_{i}^{t-1} + \vec{b}_m
\end{align}
where $\vec{s}_{i}^{t-1}$ is the state of node $i$ at time step  $t-1$, and a shared linear layer of weights $\vec{W}_m$ and biases $\vec{b}_m$ will be used to compute the messages for all nodes. After aggregating the messages, updating function of the node hidden state can be defined as follows:
\begin{align}
    \label{rgnn_state} \vec{s}_k^t & = f_{lstm} \left( \vec{r}_k^{t-1}, \vec{m}_k^t,  \vec{s}_k^{t-1} \right)
\end{align}
where $f_{lstm} \left( \cdot \right)$ denotes the LSTM cell function. Then, we calculate the relation representation $\vec{r}_k^{t}$ at time step $t$ via:
\begin{align}
    \label{rgnn_relation} \vec{r}_k^{t} & =  \vec{r}_k^{t-1} + \vec{s}_k^{t}
\end{align}
The residual design of Eqn.\ref{rgnn_relation} aims to add the relationship features between each part based on the individual part features, so that the representations contain the fusion of both features.

After the RGNN is updated $T$ times, we extract node-level output as the spatial structural relationships $\vec{r}_k^T$ of each part within each frame. Finally, the high-level spatial structural information $\vec{q}$ of human body for a frame can be computed as follows:
\begin{align}
    \label{rgnn_global1} \vec{r}^{T} & =  concat \left( [\vec{r}_1^T, \vec{r}_2^T,...,\vec{r}_k^T] \right), \forall k \in K \\
    \label{rgnn_global2} \vec{q} & =  f_r \left( \vec{r}^{T} \right)
\end{align}
where $f_{r} \left( \cdot \right)$ is a linear layer.

\subsection{Temporal Stack Learning Network}

To further exploit the discriminative features of various actions, the proposed temporal stack learning network further focus on modeling detailed temporal dynamics. For a skeleton sequence, it has rich and detailed temporal dynamics in the short-term clips. To capture the detailed temporal information, the long-term sequence can be decomposed into multiple continuous clips. In a skeleton sequence, it consists of $N$ frames. The sequence is divided into $M$ clips at intervals of $d$ frames. The high-level spatial structural features $\{Q_1, Q_2,...,Q_M \}$ of the skeleton sequence can be extracted from the spatial reasoning network. $Q_m = \{\vec{q}_{md+1}, \vec{q}_{md+2},..., \vec{q}_{(m+1)d} \}$ is the set of features of clip $m$, and $\vec{q}_n$ denotes the high-level spatial structural features of the skeleton frame $n, n \in \{1,...,N\}$.

\begin{figure}[!b]
\centering
\includegraphics[width=0.9\linewidth,height=0.4\linewidth]{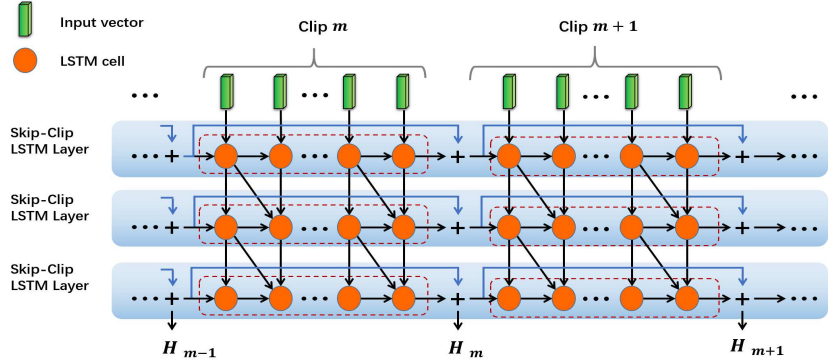}
\caption{The architecture of three skip-clip LSTM layers}
\label{clip_lstm}
\end{figure}

Our proposed temporal stack learning network is a two stream network: position network and velocity network (see Fig.~\ref{model_pipeline}). The two networks have the same architecture, which is composed of three skip-clip LSTM layers (shown in Fig.~\ref{clip_lstm}). The inputs of position network are the high-level spatial structural features $\{Q_1, Q_2,...,Q_M \}$. The inputs of velocity network are the temporal differences $\{V_1, V_2,...,V_M \}$ of the spatial features between  two consecutive frames, where $V_m = \{\vec{v}_{md+1}, \vec{v}_{md+2},..., \vec{v}_{(m+1)d} \}$. $\vec{v}_n = \vec{q}_n - \vec{q}_{n-1}$ denotes the temporal difference of high-level spatial features for the skeleton frame $n$.

\textbf{\emph{Skip-Clip LSTM Layer}} \hspace{3mm}
In the skip-clip LSTM layer, there is an LSTM layer shared among the continuous clips (see Fig.~\ref{clip_lstm}). For the position network, the spatial features of continuous skeleton frames in the clip $m$ will be fed into the shared LSTM to capture the short-term temporal dynamics in the first skip-clip LSTM layers:
\begin{align}
\label{clip_last_state}
    \vec{h}_m^{'} & =  f_{LSTM} \left( Q_m \right) \nonumber\\
    & = f_{LSTM} \left( \{\vec{q}_{md+1}, \vec{q}_{md+2},..., \vec{q}_{(m+1)d} \} \right)
\end{align}
where $\vec{h}_m^{'}$ is the last hidden state of shared LSTM for the clip $m$, $f_{LSTM} \left( \cdot \right)$ denotes the shared LSTM in the skip-clip LSTM layer.

Note that the inputs of LSTM cell between the first skip-clip LSTM layer and the other layers are different (see Fig.~\ref{clip_lstm}). In order to gain more dependency between two adjacent frames, the input $\vec{x}_t^l$ of LSTM cell for the $l$ ($l \geq 2$) layer at time step $t$ is defined as follows:
\begin{align}
\label{cell_input}
    \vec{x}_t^l & =  concat \left( \vec{h}_{t-1}^{l-1}, \vec{h}_{t}^{l-1} \right)
\end{align}
where $\vec{h}_{t}^{l-1}$ is the hidden state of the $l-1$ LSTM layer at time step $t$.

Then the representation of clip dynamics can be calculated as follows:
\begin{align}
\label{clip_state}
    \vec{H}_m & =  \vec{H}_{m-1} + \vec{h}_m^{'} \nonumber\\
    & = \sum_{i=1}^m  \vec{h}_i^{'}
\end{align}
where $\vec{H}_{m-1}$ and $\vec{H}_{m}$ denote the representations of clip $m-1$ and $m$, respectively. The representation $\vec{H}_{m}$ is to aggregate all the detailed temporal dynamics of the $m$-th clip and all previous clips to represent the long-term sequence. When feeding the clip $m$ into the shared LSTM layer, we initialize the initial hidden state $\vec{h}_m^0$ of the shared LSTM with the $\vec{H}_{m-1}$, such that $\vec{h}_m^0$ = $\vec{H}_{m-1}$, which can inherit previous dynamics to learn the short-term dynamics of the $m$-th clip to maintain the dependency between clips.

The skip-clip LSTM layer can capture the temporal dynamics of the short-term clip based on the temporal information of previous clips. And the larger $m$ is, the richer temporal dynamics $\vec{H}_m$ contains.

\textbf{\emph{Learning the Classier}} \hspace{3mm}
Finally, two linear layers are used to compute the scores for $C$ classes:
\begin{align}
\label{clip_output}
   \vec{O}_m & = F_o \left( \vec{H}_m \right)
\end{align}
where $\vec{O}_m$ is the score of clip $m$ and $\vec{O}_m = \left(o_{m1}, o_{m2},...,o_{mC}\right)$, $F_o$ denotes the two linear layers. And the output is fed to a softmax classifier to predict the probability being the $i^{th}$ class:
\begin{align}
\label{clip_softmaxt}
   {\hat y}_{mi} & = { {e^{o_{mi}}} \over { \sum_{j=1}^C e^{o_{mj}} }}, i = 1,...,C
\end{align}
where ${\hat y}_{mi}$ indicates the probability that the clip $m$ is predicted as the $i^{th}$ class. And $\vec{{\hat y}}_{m} = \left({\hat y}_{m1},...,{\hat y}_{mC} \right)$ denotes the probability vector of clip $m$.

Our proposed temporal stack learning network is a two stream network, so the clip dynamic representations ($\vec{H}_m^p$, $\vec{H}_m^v$ and $\vec{H}_m^s$) of three modes will be captured. $\vec{H}_m^p$ and $\vec{H}_m^v$ denote the dynamic representations extracted from the position and velocity for the clip $m$, respectively. And $\vec{H}_m^s$ is the sum of $\vec{H}_m^p$ and $\vec{H}_m^v$. The probability vectors ($\vec{{\hat y}}_{m}^p$, $\vec{{\hat y}}_{m}^v$ and $\vec{{\hat y}}_{m}^s$) can be predicted from the network.

In order to optimize the model, we propose the clip based incremental losses for a skeleton sequence:
\begin{align}
    \label{clip_output_p} \mathcal{L}_p  & = - \sum_{m=1}^M {m \over M} \sum_{i=1}^C  y_i log {\hat y}_{mi}^p \\
    \label{clip_output_v} \mathcal{L}_v  & = - \sum_{m=1}^M {m \over M} \sum_{i=1}^C  y_i log {\hat y}_{mi}^v \\
    \label{clip_output_s} \mathcal{L}_s  & = - \sum_{m=1}^M {m \over M} \sum_{i=1}^C  y_i log {\hat y}_{mi}^s
\end{align}
where $\vec{y} = \left( y_1,...,y_C \right)$ denotes the groundtruth label. The richer temporal information the clip contains, the greater the coefficient ${m \over M}$ is. The clip-based incremental loss will promote the ability of modeling the detailed temporal dynamics for long-term skeleton sequences. Finally, the training loss of our model is defined as follows:
\begin{align}
    \label{loss} \mathcal{L}  & = \mathcal{L}_p + \mathcal{L}_v + \mathcal{L}_s
\end{align}

Due to the mechanisms of skip-clip LSTM (see the Eqn.\ref{clip_state}), the representation $\vec{{ H}}_M^s$ of clip $M$ aggregates all the detailed temporal dynamics of the continuous clips from the position sequences and velocity sequences. In the testing process, we only use the probability vector $\vec{{\hat y}}_{M}^s$ to predict the class of the skeleton sequence.

\section{Experiments}
\label{Experiments}
To verify the effectiveness of our proposed model for skeleton-based action recognition, we perform extensive experiments on the NTU RGB+D dataset \cite{Shahroudy2016NTU} and the SYSU 3D Human-Object Interaction dataset \cite{Hu2015Jointly}. We also analyze the performance of our model with several variants.

\subsection{Datasets and Experimental Settings}
\textbf{\emph{NTU RGB+D Dataset (NTU)}} \hspace{3mm}
This is the current largest action recognition dataset with joints annotations that are collected by Microsoft Kinect v2. It has 56880 video samples and contains 60 action classes in total. These actions are performed by 40 distinct subjects. It is recorded with three cameras simultaneously in different horizontal views. The joints annotations consist of 3D locations of 25 major body joints. \cite{Shahroudy2016NTU} defines two standard evaluation protocols for this dataset: Cross-Subject and Cross-View.
For Cross-Subject evaluation, the 40 subjects are split into training and testing groups. Each group consists of 20 subjects. For Cross-View evaluation, all the samples of camera 2 and 3 are used for training while the samples of camera 1 are used for testing.

\textbf{\emph{SYSU 3D Human-Object Interaction dataset (SYSU)}} \hspace{3mm}
This dataset contains 480 video samples in 12 action classes. These actions are performed by 40 subjects. There are 20 joints for each subject in the 3D skeleton sequences. There are two standard evaluation protocols \cite{Hu2015Jointly} for this dataset. In the first setting (setting-1),  for each activity class,  half of the samples are used for training and the rest for testing. In the second setting (setting-2), half of subjects are used to train model and the rest for testing. For each setting, there is 30-fold cross validation.

\textbf{\emph{Experimental Settings}} \hspace{3mm}
In all our experiments, we set the hidden state dimension of RGNN to 256. For the NTU dataset, the human body is decomposed into $K$ = 8 parts: two arms, two hands, two legs, one trunk and one head. For the SYSU dataset, there are $K$ = 5 parts: two arms, two legs, and one trunk. We set the length $N$ = 100 of skeleton sequences for the two datasets. The neuron size of LSTM cell in the skip-clip LSTM layer is 512. The learning rate, initiated with 0.0001, is reduced by multiplying it by 0.1 every 30 epochs. The batch sizes for the NTU dataset and the SYSU dataset are 64 and 10, respectively. The network is optimized using the ADAM optimizer \cite{kingma2015adam}. Dropout with a probability of 0.5 is utilized to alleviate overfitting during training.

\subsection{Experimental Results}

We compare the performance of our proposed model against several state-of-the-art approaches on the NTU dataset and SYSU dataset in Table \ref{NTU_comparison} and Table \ref{SYSU_comparison}. These methods for skeleton-based action recognition can be divided into two categories: CNN-based methods \cite{liu2017enhanced,Yan2018Spatial} and LSTM-based methods \cite{Zhang2017View,Inwoong2017Ensemble,Song2017Attention}.

\setlength{\tabcolsep}{8pt}
\begin{table}[!t]
\fontsize{8pt}{0.85\baselineskip}\selectfont
\begin{center}
\caption{The comparison results on NTU RGB+D dataset with Cross-Subject and Cross-View settings in accuracy (\%) }
\label{NTU_comparison}
\begin{tabular}{r|cc}
\hline\noalign{\smallskip}
\multicolumn{1}{c}{Methods} & Cross-Subject & Cross-View \\
\noalign{\smallskip}
\hline
\hline
\noalign{\smallskip}
HBRNN-L \cite{Du2015Hierarchical} (2015)                 & 59.1 & 64.0 \\
Part-aware LSTM \cite{Shahroudy2016NTU} (2016)          & 62.9 & 70.3 \\
Trust Gate ST-LSTM \cite{Liu2016Spatio-temporal} (2016)  & 69.2 & 77.7 \\
Two-stream RNN \cite{Wang2017Modeling} (2017)            & 71.3 & 79.5 \\
STA-LSTM \cite{Song2017Attention} (2017)                 & 73.4 & 81.2 \\
Ensemble TS-LSTM \cite{Inwoong2017Ensemble} (2017)       & 74.6 & 81.3 \\
Visualization CNN \cite{liu2017enhanced} (2017)         & 76.0 & 82.6 \\
VA-LSTM \cite{Zhang2017View} (2017)                      & 79.4 & 87.6 \\
ST-GCN \cite{Yan2018Spatial} (2018)                      & 81.5 & 88.3 \\
\hline
SR-TSL (Ours)                                              & \textbf{84.8} & \textbf{92.4}\\
\hline
\end{tabular}
\end{center}
\end{table}

As shown in Table \ref{NTU_comparison}, we can see that our proposed model achieves the best performances of 84.8\% and 92.4\% on the current largest NTU dataset. Our performances significantly outperform the state-of-the-art CNN-based method \cite{Yan2018Spatial} by about 3.3\% and 4.1\%  for cross-subject evaluation and cross-view evaluation, respectively. Our model belongs to the LSTM-based methods. Compared with VA-LSTM \cite{Zhang2017View} that is the current best LSTM-based method for action recognition, our results are about 5.4\% and 4.8\% better than VA-LSTM on the NTU dataset. Ensemble TS-LSTM \cite{Inwoong2017Ensemble} is the most similar work to ours. The results of our model outperform by 10.2\% and 11.1\% compared with \cite{Inwoong2017Ensemble} in cross-subject evaluation and cross-view evaluation, respectively. As shown in Table \ref{SYSU_comparison}, our proposed model achieves the best performances of 80.7\% and 81.9\% on SYSU dataset, which significantly outperforms the state-of-the-art approach \cite{Zhang2017View} by about 3.8\% and 4.4\% for setting-1 and setting-2, respectively.

\setlength{\tabcolsep}{8pt}
\begin{table}[!b]
\fontsize{8pt}{0.85\baselineskip}\selectfont
\begin{center}
\caption{The comparison results on SYSU dataset in accuracy (\%) }
\label{SYSU_comparison}
\begin{tabular}{r|cc}
\hline\noalign{\smallskip}
\multicolumn{1}{c|}{Methods} & Setting-1  & Setting-2  \\
\noalign{\smallskip}
\hline
\hline
\noalign{\smallskip}
LAFF \cite{Hu2016Real} (2016)                    &   -  & 54.2 \\
Dynamic Skeletons \cite{Hu2015Jointly} (2015)    & 75.5 & 76.9 \\
VA-LSTM \cite{Zhang2017View} (2017)              & 76.9 & 77.5 \\
\hline
SR-TSL (Ours)                             & \textbf{80.7} & \textbf{81.9}\\
\hline
\end{tabular}
\end{center}
\end{table}

\subsection{Model Analysis}

We analyze the proposed model by comparing it with several baselines. The comparison results demonstrate the effectiveness of our model. There are two key ingredients in the proposed model:  spatial reasoning network (SRN) and temporal stack learning network (TSLN). To analyze the role of each component, we compare our model with several combinations of these components. Each variant is evaluated on NTU dataset.

\setlength{\tabcolsep}{8pt}
\begin{table}[!t]
\fontsize{8pt}{0.85\baselineskip}\selectfont
\begin{center}
\caption{The comparison results on NTU and SYSU dataset in accuracy (\%). We compare the performances of several variants and our proposed model to verify the effectiveness of our model}
\label{baselines_comparison}
\begin{tabular}{l|cc|cc}
    \hline\noalign{\smallskip}
    \multirow{2}{*}{Methods} & \multicolumn{2}{c|}{NTU} & \multicolumn{2}{c}{SYSU} \\
    \cline{2-5}
     &Cross-Subject & Cross-View & Setting-1  & Setting-2\\
    \noalign{\smallskip}
    \hline
    \hline
    \noalign{\smallskip}
    FC + LSTM     & 77.0   & 84.7  & 39.9  & 40.7 \\
    SRN + LSTM    & 78.7 & 87.3    & 42.1  & 44.4 \\
    FC + TSLN      & 83.8 & 91.6   & 77.3  & 77.4 \\
    SR-TSL(Position) & 78.8 & 88.2  & 77.1  & 76.9 \\
    SR-TSL(Velocity) & 82.2 & 90.6  & 71.7  & 71.8 \\
    \hline
    SR-TSL (Ours) & \textbf{84.8} & \textbf{92.4}  & \textbf{80.7}  & \textbf{81.9}\\
    \hline
\end{tabular}
\end{center}
\end{table}

\textbf{FC+LSTM}\hspace{2mm} For this model, the coordinate vectors of each body part are encoded with the linear layer and three LSTM layers are used to model the sequence dynamics. It is also a two stream network to learn the temporal dynamics from position and velocity.

\textbf{SRN+LSTM}\hspace{2mm} Compared with FC+LSTM,  this model uses spatial reasoning network to capture the high-level spatial structural features of skeleton sequences within each frame.

\textbf{FC+TSLN}\hspace{2mm} Compared with FC+LSTM, the temporal stack learning network replaces three LSTM layers to learn the detailed sequence dynamics for skeleton sequences.

\textbf{SR-TSL (Position)}\hspace{2mm} Compared with our proposed model, the temporal stack learning network of this model only contains the position network.

\textbf{SR-TSL (Velocity)}\hspace{2mm} Compared with our proposed model, the temporal stack learning network of this model only contains the velocity network.

\textbf{SR-TSL }\hspace{2mm} It denotes our proposed model.

\begin{figure}[!t]
\centering
    \subfigure[Cross-Subject]{
        \label{acc:a}
        \includegraphics[width=2.3in]{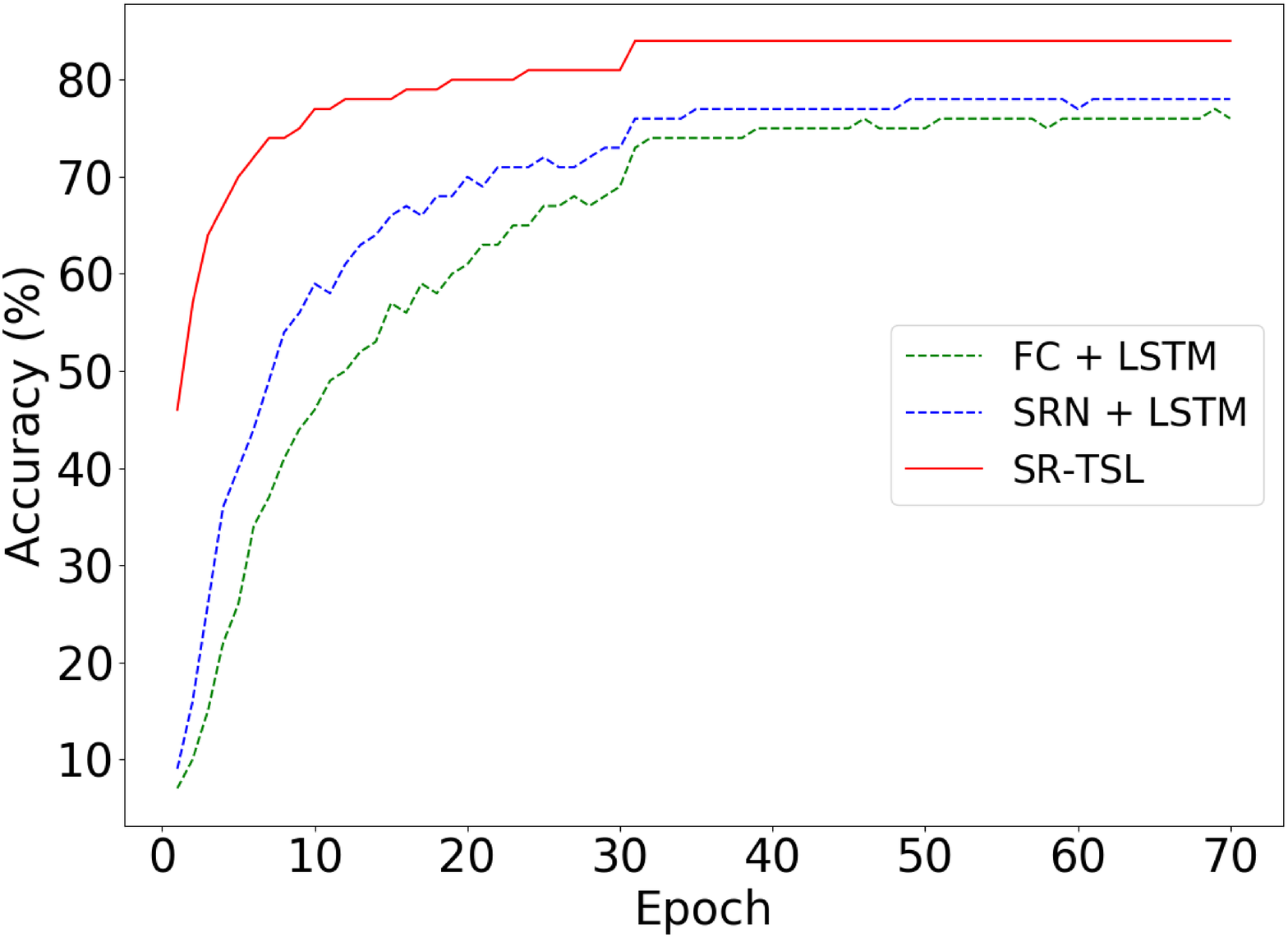}}
    \subfigure[Cross-View]{
        \label{acc:b}
        \includegraphics[width=2.3in]{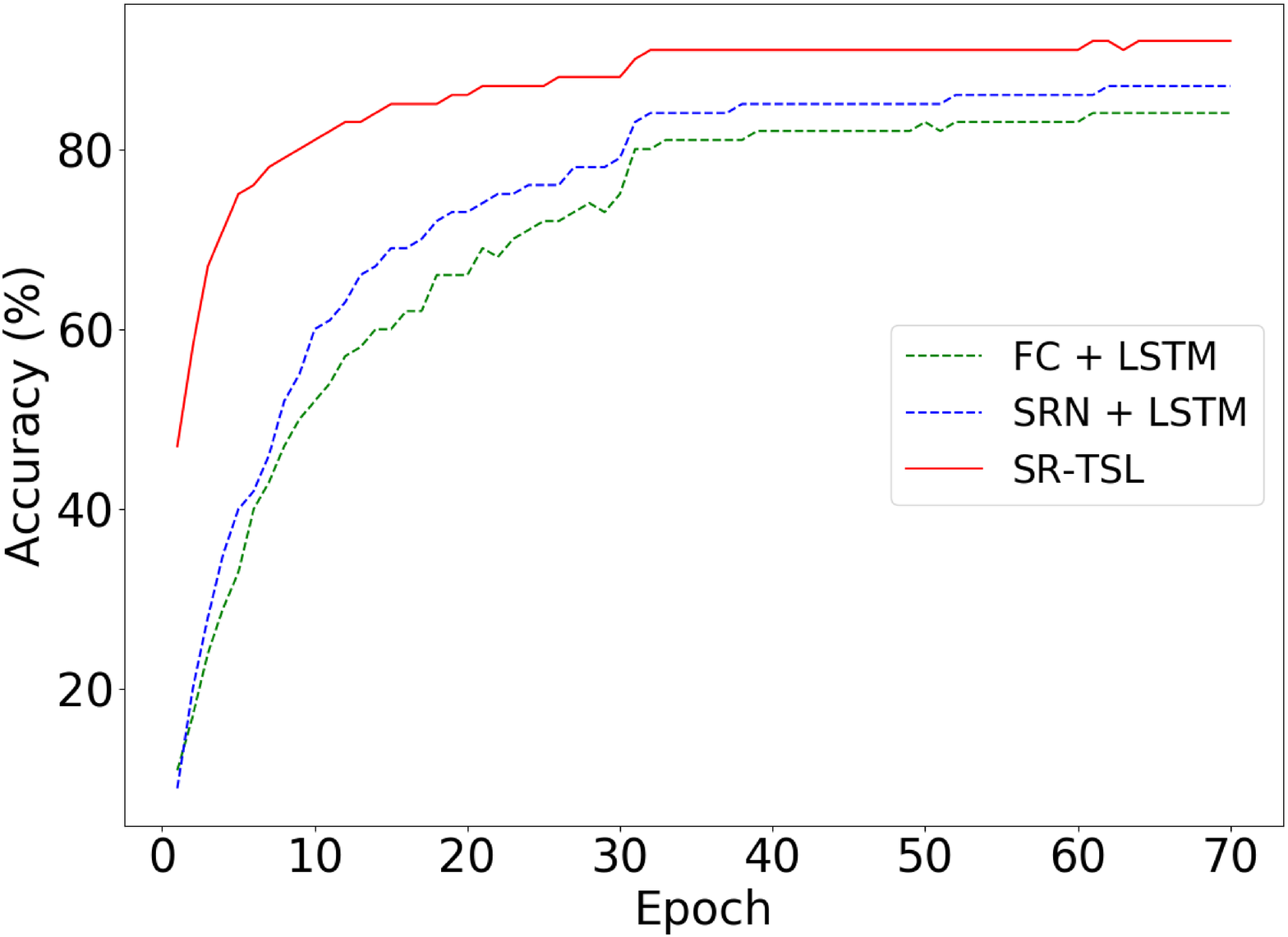}}
\caption{The accuracy of the baselines and our model on the testing set of NTU RGB+D dataset during learning phase. (a) shows the comparison results for cross-subject evaluation, and (b) is for cross-view evaluation}
\label{acc}
\end{figure}

\begin{figure}[!b]
\centering
\includegraphics[width=0.7\linewidth,height=0.38\linewidth]{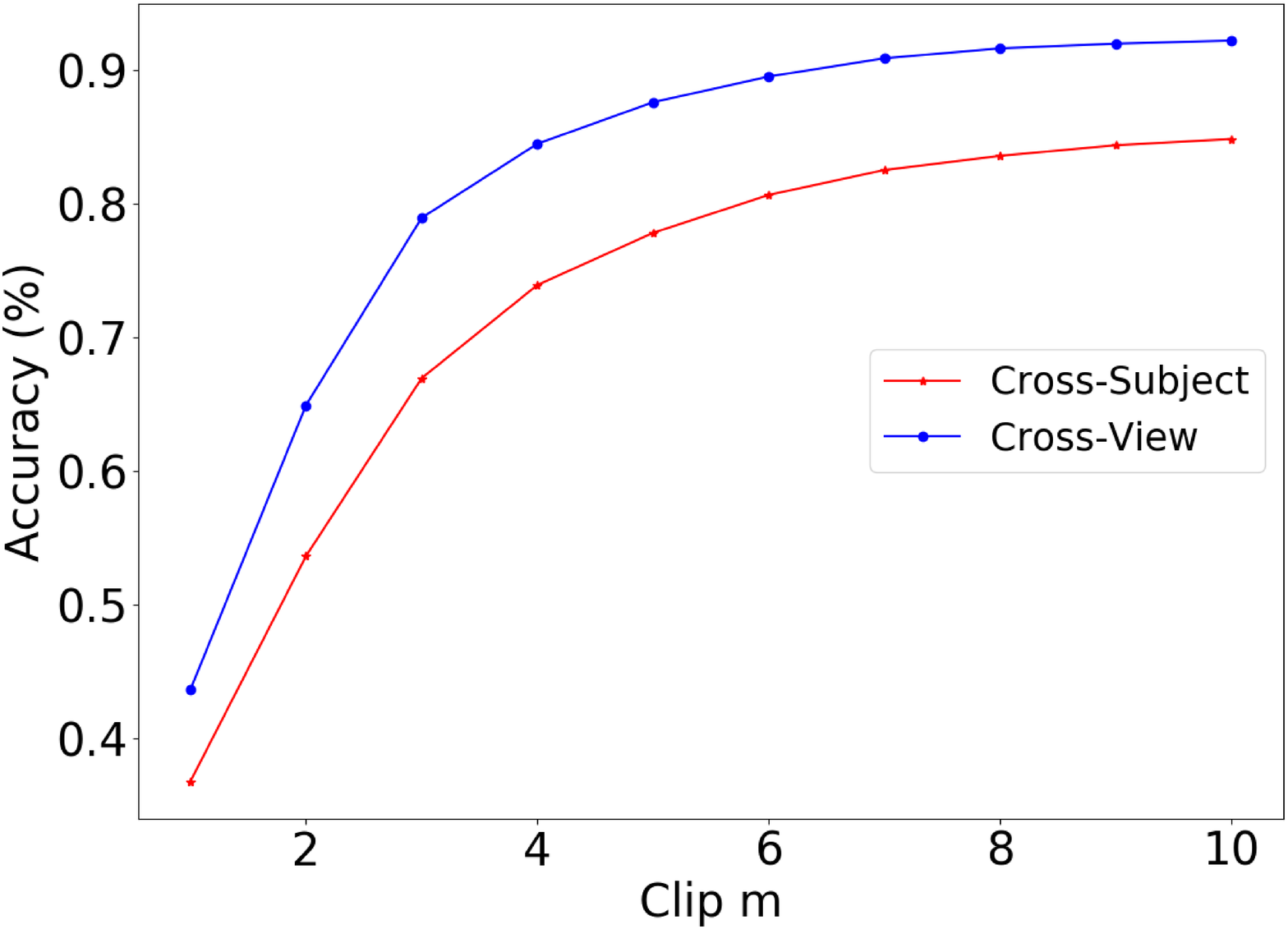}
\caption{The accuracy of the increasing clips on the testing set of NTU RGB+D dataset}
\label{clip_acc}
\end{figure}

Table \ref{baselines_comparison} shows the comparison results of the variants and our proposed model on NTU and SYSU dataset. We can observe that our model can obviously increase the performances on both datasets. And the increased performances showed in Table \ref{baselines_comparison} illustrate that the spatial reasoning network and temporal stack learning network are effective for the skeleton based action recognition, especially the temporal stack learning network. Furthermore, the two stream architecture of temporal stack learning network is efficient to learn the temporal dynamics from the velocity sequence and position sequence. Fig.~\ref{acc} shows the accuracy of the baselines and our model on the testing set of NTU RGB+D dataset during learning phase. We can see that our proposed model can speed up convergence and obviously improve the performance. We also show the process of temporal stack learning in Fig.~\ref{clip_acc}. With the increase of $m$, the much richer temporal information is contained in the representation of a sequence.
And the network can consider more temporal dynamics of the details to recognize human action, so as to improve the accuracy. The above results illustrate the proposed SR-TSL can effectively speed up convergence and obviously improve the performance.

\setlength{\tabcolsep}{6pt}
\begin{table}[t]
\begin{floatrow}

\begin{minipage}{0.4\linewidth}
\centering
\ttabbox{\caption{The comparison results on NTU dataset in accuracy (\%). We compare several models that have different time steps for the RGNN to show the improvements achieved at every step }}{%
\label{RGNN_comparison}
\begin{tabular}{l|cc}
\hline\noalign{\smallskip}
RGNN & Cross-Subject & Cross-View \\
\noalign{\smallskip}
\hline
\hline
\noalign{\smallskip}
$T$ = 1 & 84.1 & 92.0 \\
$T$ = 2 & 84.4 & 92.2 \\
$T$ = 3 & 84.5 & \textbf{92.4} \\
$T$ = 4 & 84.7 & 92.3 \\
$T$ = 5 & \textbf{84.8} & 92.3 \\
$T$ = 6 & 84.7 & 92.2 \\
\hline
\end{tabular}}
\end{minipage}

\hfil

\begin{minipage}{0.4\linewidth}
\centering
\ttabbox{\caption{The comparison results on NTU dataset in accuracy (\%). We compare the performances of several proposed models that have different the length $d$ of clips}}{%
\label{clip_comparison}
\begin{tabular}{l|cc}
\hline\noalign{\smallskip}
TSLN & Cross-Subject & Cross-View \\
\noalign{\smallskip}
\hline
\hline
\noalign{\smallskip}
$d$ = 2 & 81.6 & 90.6 \\
$d$ = 4 & 84.1 & 91.4 \\
$d$ = 6 & 84.5 & \textbf{92.4} \\
$d$ = 8 & 84.5 & 92.3 \\
$d$ = 10 & \textbf{84.8} & 92.1 \\
$d$ = 15 & 84.7 & 92.2 \\
$d$ = 20 & 84.4 & 92.1 \\
\hline
\end{tabular} }
\end{minipage}

\end{floatrow}
\end{table}

We also discuss the effect of two important hyper-parameters: the time step $T$ of the RGNN and the length $d$ of clips. The comparison results are shown in Table \ref{RGNN_comparison} and Table \ref{clip_comparison}. For the time step $T$, we can find that the performance increases by a small amount when increasing $T$, and saturates soon. We think that the high-level spatial structural features between a small number of body parts can be learned quickly. For the length $d$ of clips, with the increase of $d$, the performance is significantly improved and then saturated. The reason of saturation is that learning short-term dynamic does not require too many frames. The above experimental results illustrate that our proposed model is effective for skeleton-based action recognition.

\section{Conclusions}
\label{Conclusions}

In this paper, we propose a novel model with spatial reasoning and temporal stack learning for long-term skeleton based action recognition, which achieves much better results than the state-of-the-art methods. The spatial reasoning network can capture the high-level spatial structural information within each frame, while the temporal stack learning network can model the detailed temporal dynamics of skeleton sequences. We also propose a clip-based incremental loss to further improve the ability of stack learning, which provides an effective way to solve long-term sequence optimization. With extensive experiments on the current largest NTU RGB+D dataset and SYSU dataset, we verify the effectiveness of our model for the skeleton based action recognition. In the future, we will further analyze the error samples to improve the model, and consider more contextual information, such as interactions, to aid action recognition.

\section*{Acknowledgements}
\label{ Acknowledgements}

This work is jointly supported by National Key Research and Development Program of China (2016YFB1001000), National Natural Science Foundation of China (61525306, 61633021, 61721004, 61420106015, 61572504), Scientific Foundation of State Grid Corporation of China.

\clearpage

%
%
%
\bibliographystyle{splncs04}
\bibliography{egbib}

\begin{thebibliography}{10}
\providecommand{\url}[1]{\texttt{#1}}
\providecommand{\urlprefix}{URL }
\providecommand{\doi}[1]{https://doi.org/#1}

\bibitem{Aggarwal2011Human}
Aggarwal, J.K., Ryoo, M.S.: Human activity analysis: A review. ACM Computing
  Surveys  (2011)

\bibitem{aggarwal2014human}
Aggarwal, J.K., Xia, L.: Human activity recognition from 3d data: A review.
  Pattern Recognition Letters  (2014)

\bibitem{cao2017realtime}
Cao, Z., Simon, T., Wei, S.E., Sheikh, Y.: Realtime multi-person 2d pose
  estimation using part affinity fields. In: CVPR (2017)

\bibitem{Du2015Hierarchical}
Du, Y., Wang, W., Wang, L.: Hierarchical recurrent neural network for skeleton
  based action recognition. In: CVPR (2015)

\bibitem{NIPS2015_5954}
Duvenaud, D.K., Maclaurin, D., Iparraguirre, J., Bombarell, R., Hirzel, T.,
  Aspuru-Guzik, A., Adams, R.P.: Convolutional networks on graphs for learning
  molecular fingerprints. In: NIPS (2015)

\bibitem{henaff2015deep}
Henaff, M., Bruna, J., LeCun, Y.: Deep convolutional networks on
  graph-structured data. arXiv preprint arXiv:1506.05163  (2015)

\bibitem{Hochreiter1997lstm}
Hochreiter, S., Schmidhuber, J.: Long short-term memory. Neural Computation
  (1997)

\bibitem{Hu2015Jointly}
Hu, J.F., Zheng, W.S., Lai, J., Zhang, J.: Jointly learning heterogeneous
  features for rgb-d activity recognition. In: CVPR (2015)

\bibitem{Hu2016Real}
Hu, J.F., Zheng, W.S., Ma, L., Wang, G., Lai, J.: Real-time rgb-d activity
  prediction by soft regression. In: ECCV (2016)

\bibitem{Hussein2013Human}
Hussein, M.E., Torki, M., Gowayyed, M.A., El-Saban, M.: Human action
  recognition using a temporal hierarchy of covariance descriptors on 3d joint
  locations. In: IJCAI (2013)

\bibitem{johansson1973visual}
Johansson, G.: Visual perception of biological motion and a model for its
  analysis. Perception \& psychophysics  (1973)

\bibitem{Ke2017A}
Ke, Q., Bennamoun, M., An, S., Sohel, F., Boussaid, F.: A new representation of
  skeleton sequences for 3d action recognition. In: CVPR (2017)

\bibitem{Kim2017Interpretable}
Kim, T.S., Reiter, A.: Interpretable 3d human action analysis with temporal
  convolutional networks. In: CVPR Workshops (2017)

\bibitem{kingma2015adam}
Kingma, D.P., Ba, J.: Adam: A method for stochastic optimization. In: ICLR
  (2015)

\bibitem{Bruna2014Spectral}
LeCun, J.B.W.Z.A.S.Y.: Spectral networks and locally connected networks on
  graphs. In: ICLR (2014)

\bibitem{Inwoong2017Ensemble}
Lee, I., Kim, D., Kang, S., Lee, S.: Ensemble deep learning for skeleton-based
  action recognition using temporal sliding lstm networks. In: ICCV (2017)

\bibitem{Li_2017_ICCV}
Li, R., Tapaswi, M., Liao, R., Jia, J., Urtasun, R., Fidler, S.: Situation
  recognition with graph neural networks. In: ICCV (2017)

\bibitem{Liu2016Spatio-temporal}
Liu, J., Shahroudy, A., Xu, D., Wang, G.: Spatio-temporal lstm with trust gates
  for 3d human action recognition. In: ECCV (2016)

\bibitem{liu2017enhanced}
Liu, M., Liu, H., Chen, C.: Enhanced skeleton visualization for view invariant
  human action recognition. Pattern Recognition  (2017)

\bibitem{niepert2016learning}
Niepert, M., Ahmed, M., Kutzkov, K.: Learning convolutional neural networks for
  graphs. In: ICML (2016)

\bibitem{Poppe2010survey}
Poppe, R.: A survey on vision-based human action recognition. Image and vision
  computing  (2010)

\bibitem{Qi_2017_ICCV}
Qi, X., Liao, R., Jia, J., Fidler, S., Urtasun, R.: 3d graph neural networks
  for rgbd semantic segmentation. In: ICCV (2017)

\bibitem{Scarselli2009graph}
Scarselli, F., Gori, M., Tsoi, A.C., Hagenbuchner, M., Monfardini, G.: The
  graph neural network model. IEEE Transactions on Neural Networks  (2009)

\bibitem{Shahroudy2016NTU}
Shahroudy, A., Liu, J., Ng, T.T., Wang, G.: Ntu rgb+d: A large scale dataset
  for 3d human activity analysis. In: CVPR (2016)

\bibitem{Simonyan2014Two-stream}
Simonyan, K., Zisserman, A.: Two-stream convolutional networks for action
  recognition in videos. In: NIPS (2014)

\bibitem{Song2017Attention}
Song, S., Lan, C., Xing, J., Zeng, W., Liu, J.: An end-to-end spatio-temporal
  attention model for human action recognition from skeleton data. In: AAAI
  (2017)

\bibitem{Raviteja2014Human}
Vemulapalli, R., Arrate, F., Chellappa, R.: Human action recognition by
  representing 3d skeletons as points in a lie group. In: CVPR (2014)

\bibitem{Wang2017Modeling}
Wang, H., Wang, L.: Modeling temporal dynamics and spatial configurations of
  actions using two-stream recurrent neural networks. In: CVPR (2017)

\bibitem{Wang2012Mining}
Wang, J., Liu, Z., Wu, Y., Yuan, J.: Mining actionlet ensemble for action
  recognition with depth cameras. In: CVPR (2012)

\bibitem{Weinland2011survey}
Weinland, D., Ronfard, R., Boyer, E.: A survey of vision-based methods for
  action representation, segmentation and recognition. Computer vision and
  image understanding  (2011)

\bibitem{Yan2018Spatial}
Yan, S., Xiong, Y., Lin, D., xiaoou Tang: Spatial temporal graph convolutional
  networks for skeleton-based action recognition. In: AAAI (2018)

\bibitem{Zhang2017View}
Zhang, P., Lan, C., Xing, J., Zeng, W., Xue, J., Zheng, N.: View adaptive
  recurrent neural networks for high performance human action recognition from
  skeleton data. In: ICCV (2017)

\bibitem{zhang2012microsoft}
Zhang, Z.: Microsoft kinect sensor and its effect. IEEE multimedia  (2012)

\bibitem{Zhu2016Co-Occurrence}
Zhu, W., Lan, C., Xing, J., Zeng, W., Li, Y., Shen, L., Xie, X.: Co-occurrence
  feature learning for skeleton based action recognition using regularized deep
  lstm networks. In: AAAI (2016)

\end{thebibliography}
%
%
%
%
%
\end{document}